%% file: main.tex
\def\year{2023}\relax
\documentclass[letterpaper]{article} 
\usepackage{aaai23}  
\usepackage{times}  
\usepackage{helvet}  
\usepackage{courier}  
\usepackage[hyphens]{url}  
\usepackage{graphicx} 
\usepackage{natbib}  
\usepackage{caption} 
\DeclareCaptionStyle{ruled}{labelfont=normalfont,labelsep=colon,strut=off} 
\usepackage{algorithm}
\usepackage{placeins}
\usepackage{algorithmicx}
\usepackage{tabularx}
\usepackage{graphicx}
\usepackage{placeins}
\usepackage{centernot}
\usepackage{siunitx}
\usepackage{xspace}
\usepackage{amsmath}
\usepackage{hyperref}
\usepackage{booktabs}
\usepackage{multirow}
\usepackage{subcaption}
\usepackage{makecell}
\usepackage{svg}
\usepackage[noend]{algpseudocode}
\newtheorem{theorem}{Theorem}

\newenvironment{proofsketch}{{\bfseries Proof sketch}}{}

\usepackage{newfloat}
\usepackage{listings}
\floatstyle{ruled}
\newfloat{listing}{tb}{lst}{}
\floatname{listing}{Listing}
\title{A-ePA*SE: Anytime Edge-Based Parallel A* for Slow Evaluations}

\iftrue
\author{
Hanlan Yang, Shohin Mukherjee, Maxim Likhachev
}
\affiliations{
    The Robotics Institute, CMU\\
    \{hanlany, shohinm, mlikhach\}@andrew.cmu.edu\\
}
\fi

\newcommand{\pase}{PA*SE\xspace}
\newcommand{\epase}{ePA*SE\xspace}
\newcommand{\gepase}{G\epase\xspace}

\newcommand{\astar}{A*\xspace}
\newcommand{\wastar}{w\astar}
\newcommand{\arastar}{ARA*\xspace}
\newcommand{\wepase}{\mbox{w-}\epase}

\newcommand{\aepase}{A-\epase}
\newcommand{\aepasenaive}{\aepase-naive\xspace}
\newcommand{\state}{\ensuremath{\mathbf{s}}\xspace}

\newcommand{\States}{\ensuremath{\mathcal{S}}\xspace}
\newcommand{\startstate}{\ensuremath{\state_0}\xspace}
\newcommand{\goalstate}{\ensuremath{\state_g}\xspace}
\newcommand{\ac}{\ensuremath{\mathbf{a}}\xspace}
\newcommand{\dac}{\ensuremath{\mathbf{a^d}}\xspace}
\newcommand{\pac}{\ensuremath{\mathbf{a^d}}\xspace}
\newcommand{\Aset}{\ensuremath{\mathcal{A}}\xspace}
\newcommand{\goalreg}{\ensuremath{\mathcal{G}}\xspace}
\newcommand{\ed}{\ensuremath{e}\xspace}

\newcommand{\edge}{\ensuremath{(\mathbf{s},\mathbf{a})}\xspace}

\newcommand{\pedgenext}{\ensuremath{(\mathbf{s}',\mathbf{a^d})}\xspace}
\newcommand{\open}{\ensuremath{\textit{OPEN}}\xspace}
\newcommand{\incon}{\ensuremath{\textit{INCON}}\xspace}

\newcommand{\closed}{\ensuremath{\textit{CLOSED}}\xspace}
\newcommand{\be}{\ensuremath{\textit{BE}}\xspace}
\newcommand{\gval}{\ensuremath{g}}

\newcommand{\gopt}{\gval^*}
\newcommand{\cost}{\ensuremath{c}\xspace}
\newcommand{\costopt}{\ensuremath{\cost^*}\xspace}
\newcommand{\fval}{\ensuremath{f}}
\newcommand{\hval}{\ensuremath{h}}
\newcommand{\wh}{\ensuremath{w}\xspace}
\newcommand{\wi}{\ensuremath{\epsilon}\xspace}
\newcommand{\numexpanded}{\ensuremath{n\_successors\_generated}\xspace}
\newcommand{\graph}{\ensuremath{G}\xspace}
\newcommand{\vertex}{\ensuremath{v}\xspace}
\newcommand{\Vertices}{\ensuremath{\mathcal{V}}\xspace}
\newcommand{\Edges}{\ensuremath{\mathcal{E}}\xspace}
\newcommand{\plan}{\ensuremath{\pi}\xspace}
\newcommand{\numthreads}{\ensuremath{N_t}\xspace}
\newcommand{\timebudget}{\ensuremath{T}\xspace}
\newcommand{\speedup}{\ensuremath{s}\xspace}

\newcommand{\improve}{\textsc{ImprovePath}\xspace}

\begin{document}

\maketitle

\begin{abstract}
\input{01abstract}
\end{abstract}

\FloatBarrier
\section{Introduction}
\input{02introduction}

\FloatBarrier
\section{Related Work}
\input{03background}

\FloatBarrier
\section{Problem Definition}
\label{sec:problem}
\input{04problem}

\FloatBarrier
\section{Method}
\label{sec:methods}
\input{05methods}

\FloatBarrier
\section{Properties}
\label{sec:Properties}

\input{06properties}

\FloatBarrier
\section{Evaluation}
\input{07evaluation}

\FloatBarrier
\section{Conclusion and Future Work}
\input{08conclusion_future}

\FloatBarrier
\section{Acknowledgements}
\input{09acknowledgements}

\FloatBarrier
\bibliography{main}

\end{document}

%% file: 01abstract.tex
Anytime search algorithms are useful for planning problems where a solution is desired under a limited time budget. Anytime algorithms first aim to provide a feasible solution quickly and then attempt to improve it until the time budget expires. On the other hand, parallel search algorithms utilize the multithreading capability of modern processors to speed up the search. One such algorithm, \epase (Edge-Based Parallel A* for Slow Evaluations), parallelizes edge evaluations to achieve faster planning and is especially useful in domains with expensive-to-compute edges. In this work, we propose an extension that brings the anytime property to \epase, resulting in \aepase. We evaluate \aepase experimentally and show that it is significantly more efficient than other anytime search methods. The open-source code for \aepase along with the baselines is available here: \url{https://github.com/shohinm/parallel_search}

%% file: 02introduction.tex
Graph search algorithms are widely used in robotics for planning which can be formulated as the shortest path problem on a graph~\cite{kusnur2021planning, mukherjee2021reactive}. Parallelized graph search algorithms have shown to be effective in robotics domains where action evaluation tends to be expensive. In particular, a parallelized planning algorithm \epase (Edge-based Parallel A* for Slow Evaluations) was developed~\cite{mukherjee2022epase} that changes the basic unit of the search from state expansions to edge expansions. This decouples the evaluation of edges from the expansion of their common parent state, giving the search the flexibility to figure out what edges need to be evaluated to solve the planning problem. Additionally, this provides a framework for the asynchronous parallelization of edge evaluations within the search.

Though parallelized planning algorithms achieve drastically lower planning times than their serial counterparts, for their applicability in real-time robotics, planning algorithms need to come up with a solution under a strict time budget. Though the optimal solution is preferable, that is often not the first priority. For such domains, anytime algorithms have been developed that first prioritize a quick feasible solution by allowing a high sub-optimality bound. This is typically done by incorporating a high inflation factor on the heuristic. They then attempt to improve the solution by incrementally decreasing the inflation factor and therefore tightening the sub-optimality bound until the time runs out. Therefore in this work, we bring the anytime property to \epase. We show that the resulting algorithm, \aepase, achieves higher efficiency than existing anytime algorithms. 

%% file: 03background.tex
\subsubsection{Anytime algorithms}

A naive approach to make \wastar anytime is to sequentially run several iterations of it from scratch while reducing the heuristic inflation. Anytime A* \cite{anytimeAstar} finds an initial solution using \wastar and then continues searching to improve the solution. A more elegant anytime algorithm Anytime Repairing A* (\arastar) \cite{likhachev2003ara} reuses previous search efforts to prevent redundant work by keeping track of states whose cost-to-come can be further reduced in future iterations. Anytime Multi-heuristic A* (A-MHA*) \cite{natarajan2019amha}  brings the anytime property to Multi-heuristic A* \cite{aine2016multi}. Anytime Multi-resolution Multi-heuristic A* (AMRA*) \cite{saxena2022amra} is an anytime algorithm that searches over multiple resolutions of the state space. These algorithms, however, do not utilize any parallelization.

\subsubsection{Parallel algorithms}
Sampling-based methods like PRMs can be trivially parallelized~\cite{amato1999probabilistic} by utilizing parallel processes cooperatively build the roadmap~\cite{jacobs2012scalable}. Parallelized versions of RRT also exist in which multiple processes expand the search tree by sampling and adding multiple new states in parallel~\cite{devaurs2011parallelizing, ichnowski2012parallel, jacobs2013scalable, park2016parallel}. However, in many planning domains, sampling of states is not trivial, like in the case of domains that use a simulator in the loop~\cite{liang2021search}. Parallelizing search-based methods are non-trivial because of their sequential nature. However, there have been several algorithms developed that achieve this. Parallel A*~\cite{irani1986parallel} expands states in parallel while allowing re-expansions to maintain optimality, resulting in a high number of expansions. Several other approaches that parallelize state expansions suffer from this downside~\cite{evett1995massively, zhou2015massively, burns2010best}, especially if they employ a weighted heuristic. In contrast, \pase~\cite{phillips2014pa} parallelly expands states at most once, in a way that does not affect the bounds on the solution quality. \epase~\cite{mukherjee2022epase} improves \pase by changing the basic unit of the search from state expansions to edge expansions and then parallelizing this search over edges. \gepase~\cite{mukherjee2023gepase} extends \epase to domains where the actions are heterogenous in computational effort. A parallelized lazy planning algorithm, MPLP~\cite{mukherjee2022mplp}, achieves faster planning by running the search and evaluating edges asynchronously in parallel. There has also been work on parallelizing A* search on GPUs~\cite{zhou2015massively, {he2021efficient}} or multiple GPUs~\cite{he2021efficient} by utilizing multiple parallel priority queues. These algorithms have a fundamental limitation that stems from the SIMD (single-instruction-multiple-data) execution model of a GPU, which limits their applicability to domains with simple actions that share the same code.

%% file: 04problem.tex
Let a finite graph $\graph = (\Vertices, \Edges)$ be defined as a set of vertices \Vertices and directed edges \Edges. Each vertex $\vertex \in \Vertices$ represents a state \state in the state space of the domain \States. An edge $\ed \in \Edges$ connecting two vertices $\vertex_1$ and $\vertex_2$ in the graph represents an action $\ac \in \Aset$ that takes the agent from corresponding states $\state_1$ to $\state_2$. In this work, we assume that all actions are deterministic. Hence an edge \ed can be represented as a pair \edge, where \state is the state at which action \ac is executed. For an edge \ed, we will refer to the corresponding state and action as $\ed.\state$ and $\ed.\ac$ respectively. \startstate is the start state and \goalreg is the goal region. $\cost:\Edges \rightarrow [0,\infty]$ is the cost associated with an edge. $\gval(\state)$ or g-value is the cost of the best path to \state from \startstate found by the algorithm so far and $\hval(\state)$ is a consistent heuristic~\cite{russell2010artificial}. Additionally, there exists a forward-backward consistent~\cite{phillips2014pa} pairwise heuristic function $\hval(\state, \state')$ that provides an estimate of the cost between any pair of states. A path \plan is defined by an ordered sequence of edges $\ed_{i=1}^N = \edge_{i=1}^N$, the cost of which is denoted as  $\cost(\plan) = \sum_{i=1}^N \cost(\ed_i)$. The objective is to find a path \plan from $\state_0$ to a state in the goal region \goalreg within a time budget \timebudget. There is a computational budget of \numthreads threads available, which can run in parallel.

%% file: 05methods.tex
We first describe \epase and then the anytime extension to get to \aepase.

\subsubsection{ePA*SE}

\input{aepase.tex}

In A*, during a state expansion, all its successors are generated and are inserted/repositioned in the open list. In \epase, the open list (\open) is a priority queue of edges (not states) that the search has generated but not expanded, where the edge with the smallest key/priority is placed in the front of the queue. The priority of an edge $\ed=\edge$ in \open is $\fval\left(\edge\right) = \gval(\state) + \hval(\state)$. Expansion of an edge \edge involves evaluating the edge to generate the successor $\state'$ and adding/updating (but not evaluating) the edges originating from $\state'$ into \open with the same priority of $\gval(\state') + \hval(\state')$. Henceforth, whenever $\gval(\state')$ changes, the positions of all of the outgoing edges from $\state'$ need to be updated in \open. To avoid this, \epase replaces all the outgoing edges from $\state'$ by a single \textit{dummy} edge $(\state', \dac)$, where \pac stands for a dummy action until the dummy edge is expanded. Every time $\gval(\state')$ changes, only the dummy edge has to be repositioned. Unlike what happens when a real edge is expanded, when the dummy edge $(\state', \dac)$ is expanded, it is replaced by the outgoing real edges from $\state'$ in \open. This is also when the state $\state'$ is considered to be under expansion. The real edges are expanded when they are popped from \open by an edge expansion thread. This means that every edge gets delegated to a separate thread for expansion. $\state'$ is marked expanded (Line~\ref{alg:aepase_expand/add_closed}, Alg. ~\ref{alg:aepase_expand}) when all outgoing edges are expanded.

A single thread runs the main planning loop (Alg. \ref{alg:aepase_improvepath}) and pulls out edges from \open, and delegates their expansion to an edge expansion thread (Alg.\ref{alg:aepase_expand}). To maintain optimality, an edge can only be expanded if it is independent of all edges ahead of it in \open and the edges currently being expanded, i.e., in set \be~\cite{mukherjee2022epase}. An edge \ed is independent of another edge $\ed'$ if the expansion of $\ed'$ cannot possibly reduce $\gval(\ed.\state)$. Formally, this independence check is expressed by Inequalities~\ref{eq:ind_check_1}~and~\ref{eq:ind_check_2}. \wepase is a bounded suboptimal variant of \epase that trades off optimality for faster planning by introducing two inflation factors. $\wh\geq1$ inflates the priority of edges in \open i.e. $\fval\left(\edge\right)=\gval(\state)~+~\wh\hval(\state)$. $\wi\geq1$ used in Inequalities \ref{eq:ind_check_1} and ~\ref{eq:ind_check_2} relaxes the independence rule. As long as $\wi\geq\wh$, the solution cost is bounded by $\wi\cdot\costopt$. We let $\wi = \wh$ in this work, so we have one variable to control the suboptimality bound.

\begin{align}
    \label{eq:ind_check_1}
    \begin{split}
        \gval(\ed.\state) - \gval(\ed'.\state) \leq \wi\hval(\ed'.\state, \ed.\state)\\
        \forall\ed' \in  \open~|~\fval\left(\ed'\right) < \fval\left(\ed\right)
    \end{split}
\end{align}

\begin{align}
    \label{eq:ind_check_2}
    \begin{split}
        \gval(\ed.\state) - \gval(\state') \leq \wi\hval(\state', \ed.\state)~\forall\state' \in \be
    \end{split}
\end{align}

\subsubsection{\aepase}

Inspired by \arastar, we extend \wepase to a parallelized anytime repairing algorithm \aepase by inheriting three algorithmic techniques:
\begin{enumerate}
    \item Define  locally inconsistent states as the states whose \gval-values change while they are in $\closed \cup \be$ during the current \improve execution (\cite{likhachev2003ara}). \aepase keeps track of locally inconsistent states by maintaining an inconsistent list \incon (Line~\ref{alg:aepase_expand/add_incon}, Alg.~\ref{alg:aepase_expand}).
    
    \item After every $i^{th}$ \improve call exits, \aepase initializes \open for the next search iteration $i+1$ as $\open_{i+1} = \open_i \cup \incon$ . (Line \ref{alg:aepase/init_open}, Alg.\ref{alg:aepase_plan_loop}).
    
    \item \aepase changes the termination condition of a search iteration (Line~\ref{alg:aepase/termination_check}, Alg.\ref{alg:aepase_improvepath}) so that it only expands states that 1) have \gval-value that can be lowered in the current \improve iteration or 2) were locally inconsistent in the previous \improve iteration.
\end{enumerate}

\aepase extends \wepase with an additional outer control loop (Alg.~\ref{alg:aepase_plan_loop}) that sequentially reduces \wh. In the first iteration, \improve is called with $\wh_0$. This is equivalent to running \wepase except for the algorithmic change described in technique 1. When \improve returns, the current \wh-suboptimal solution is published (Line \ref{alg:aepase/publish_sol}, Alg.~\ref{alg:aepase_plan_loop}). Before every subsequent call to \improve, \wh is reduced by $\Delta\wh$ and \open is updated as described in technique 2. It is possible that no or very few states in  \open satisfy the termination check stated in technique 3 and \improve returns right away or after a few expansions. This reusing of previous search effort is the fundamental source of efficiency gains for \aepase as compared to running \wepase from scratch with a reduced \wh. \aepase terminates when either 1) the time budget expires, and the current best solution is returned or 2) \improve finds a provably optimal solution with $\wh = 1$.

%% file: aepase.tex
\begin{algorithm}[b]
\caption{\label{alg:aepase_plan_loop} \aepase : Plan}
\begin{footnotesize}
\begin{algorithmic}[1]
\State $\Aset\gets\text{action space }$, $\numthreads\gets$ thread budget, $\timebudget\gets$ time budget
\State $\wh_0\gets\text{initial heuristic weight}$, $\Delta \wh\gets\text{delta heuristic weight }$
\State $\graph \gets \text{graph}$, $\startstate\gets\text{start state }$, $\goalreg\gets\text{goal region}$
\State $terminate \gets \text{False}$
\Procedure{Plan}{}
    \State $\incon\gets\emptyset$
    \State $\forall\state\in\graph$,~$\state.\gval\gets\infty$
    \State $\state_0.\gval\gets0$, $\wh = \wh_0$
    \State insert $(\startstate, \dac)$ in \open
    \Comment{Dummy edge from \startstate}
    \While{$\wh >= 1 $ \textbf{and not} $\textsc{Timeout}\left(\timebudget\right)$}
    \label{alg:aepase/main_termination}
        \State $\incon=\emptyset, \closed=\emptyset$
        \State $\textsc{ImprovePath}(\wh)$
        \label{alg:aepase/call_improve}
        \State $\text{Publish current $\wh$ bounded sub-optimal solution}$
        \label{alg:aepase/publish_sol}
        \State $\wh = \wh - \Delta \wh$
        \State $\open=\open\cup\incon$
        \label{alg:aepase/init_open}
        \State $\text{Re-balance \open with new \wh}$
    \EndWhile 
    \State $terminate = \text{True}$
\EndProcedure
\end{algorithmic}
\end{footnotesize}
\end{algorithm}

\begin{algorithm}[t]
\caption{\label{alg:aepase_improvepath} \aepase: ImprovePath}
\begin{footnotesize}
\begin{algorithmic}[1]
\Procedure{ImprovePath}{}
    \State LOCK
    \While{$\fval(\goalstate) > min_{\state\in\open}(\fval(\state))$}
    \label{alg:aepase/termination_check}
        \If{$\open=\emptyset\textbf{ and }\be=\emptyset$}
            \State UNLOCK
            \State $\Return~\emptyset$
        \EndIf
        \State remove an edge \edge from \open that has the \newline\hspace*{2.9em} smallest $\fval(\edge)$ among all states in \open that \newline\hspace*{2.9em} satisfy Equations \ref{eq:ind_check_1} and \ref{eq:ind_check_2}
        \label{alg:aepase/open_pop}
        \If{such an edge does not exist}
            \State UNLOCK
            \State wait until \open or \be change
            \label{alg:aepase/wait}
            \State LOCK
            \State continue
        \EndIf
        \If{$\state \in \goalreg$ \textbf{and} $\fval(\goalstate) > \fval(\state)$}
            \State $\goalstate = \state$
            \State $plan = \textsc{Backtrack(\state)}$
            \label{alg:aepase/construct_path}
        \EndIf
        \State UNLOCK
        \While{\edge has not been assigned a thread}
        \label{alg:aepase/assign_start}
            \For{$i=1:\numthreads$}
                \If{thread $i$ is available}
                    \If{thread $i$ has not been spawned}
                        \State Spawn $\textsc{EdgeExpandThread}(i)$
                        \label{alg:aepase/spawn}
                    \EndIf
                    \State Assign \edge to thread $i$
                    \label{alg:aepase/assign_edge}
                \EndIf
            \EndFor
        \EndWhile
        \State LOCK
    \EndWhile 
    \State UNLOCK
    \State $\Return~plan$
\EndProcedure
\end{algorithmic}
\end{footnotesize}
\end{algorithm}

\begin{algorithm}[t]
\caption{\label{alg:aepase_expand} \aepase: Edge Expansion}
\begin{footnotesize}
\begin{algorithmic}[1]
\Procedure{EdgeExpandThread}{$i$}
    \While{$\textbf{not } terminate$}
        \If{thread $i$ has been assigned an edge \edge}
            \State $\textsc{Expand}\left(\edge\right)$
        \EndIf
    \EndWhile
\EndProcedure
\Procedure{Expand}{$\edge$}
    \State LOCK
    \If{$\ac = \pac$}
        \State insert \state in \be
        \label{alg:aepase_expand/add_be}
        \For{$\ac \in \Aset$}
            \State $\fval\left(\edge\right) = \gval(\state) + \hval(\state)$
            \State insert \edge in \open with $\fval\left(\edge\right)$
            \label{alg:aepase_expand/add_realedge}
        \EndFor
    \Else
        \State UNLOCK
        \If{ $\textsc{NotEvaluated}\left(\edge\right)$ }
            \State $\state',  \cost\left(\edge\right) \gets \textsc{GenerateSuccessor}
            \left(\edge\right)$
        \Else
            \State $\state',  \cost\left(\edge\right) \gets \textsc{GetSuccessor}
            \left(\edge\right)$
        \EndIf
        \label{alg:aepase_expand/evaluate}
        \State LOCK
        \If{$\gval (\state')>\gval(\state)+\cost\left(\edge\right)$}
            \State $\gval(\state') = \gval(\state) + \cost\left(\edge\right)$
            \State $\fval\left(\pedgenext\right) = \gval(\state') + \wh\hval(\state')$
            \If{$\state' \notin \closed\cup\be$}
                \State update \pedgenext in \open with $\fval\left(\pedgenext\right)$
            \Else
                \State update \pedgenext in \incon with $\fval\left(\pedgenext\right)$
                \label{alg:aepase_expand/add_incon}
            \EndIf
        \EndIf
        \State $\numexpanded(\state)+=1$
        \If{$\numexpanded(\state) = |\Aset|$}
            \State remove \state from \be
            \label{alg:aepase_expand/remove_be}
            \State insert \state in \closed
            \label{alg:aepase_expand/add_closed}
        \EndIf
    \EndIf
    \State UNLOCK
\EndProcedure
\end{algorithmic}
\end{footnotesize}
\end{algorithm}

%% file: 06properties.tex
\begin{theorem}
    (\textbf{Anytime correctness}) Each time the \improve function exits, the following holds: the cost of a greedy path from \startstate to \goalstate is no larger than $\lambda\gopt(\goalstate)$, where $\lambda = \max(\wi, \wh)$.
\end{theorem}
\begin{theorem}
    (\textbf{Anytime efficiency}) Within each call to \improve a state \state is expanded only if it was already locally inconsistent before the call to \improve or its \gval-value was lowered during the current execution of \improve.
\end{theorem}
\begin{proofsketch}
  These properties were proved for each \improve function call in \arastar (Corollary 13 \& Theorem 2 in \cite{likhachev2003ara-proof}) with $\lambda=\wh$. The anytime correctness properties are also proved for a single \wepase run (Theorem 3 in \cite{mukherjee2022epase}) with $\lambda = \max(\wi, \wh)$. Since we are inheriting the method to repair the graph and reuse the search effort of \arastar, these properties similarly follow for each \improve function call in \aepase.
\end{proofsketch}

%% file: 07evaluation.tex
\begin{figure}[!htb]
    \centering
    \includegraphics[width=\columnwidth]{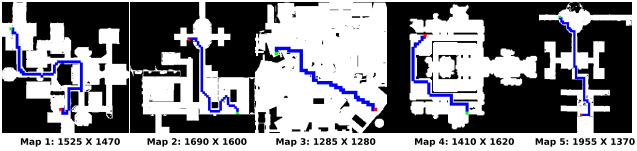}
    \caption{MovingAI maps with the start state shown in green, the goal state shown in red, and the computed path shown in blue.}
    \label{aepase/fig/2d_maps}
\end{figure}

We use 5 scaled MovingAI 2D maps~\cite{sturtevant2012benchmarks}, with state space being 2D grid coordinates shown in Fig.~\ref{aepase/fig/2d_maps}. The agent has a square footprint with a side length of 32 units. The action space comprises moving along 8 directions by 25 cell units. To check action feasibility, we collision-check the footprint at interpolated states with a 1-unit discretization. For each map, we sample 50 random start-goal pairs and verify that there exists a solution by running \wastar with a large timeout. All algorithms use Euclidean distance as the heuristic. We run the experiments with two cost maps: 1) Euclidean cost and 2) Euclidean cost multiplied with a random factor map generated by sampling a uniform distribution between 1 and 100. In the random cost map, there is a tendency for the solution to be improved more gradually with the decrease in \wh. In the case of Euclidean cost, the solution tends to improve only from one topology to another with a \wh decrease, yielding fewer intermediate suboptimal solutions. We compare \aepase with three baselines: 1) \arastar 2) \epase and 3) \aepasenaive, which runs \wepase sequentially with decreasing \wh without reuse of previous search effort. For the anytime algorithms, $\wh_0$ is set to 50, and $\Delta\wh$ is set to 0.5. All experiments were carried out on an AMD Threadripper Pro 5995WX workstation with a thread budget of 120. In all cases, we keep a high time budget, so none of the algorithms timeout.

\begin{table}[!hb]
\footnotesize
\centering
\resizebox{\columnwidth}{!}{%
\begin{tabular}{c|ccc|ccc}
\toprule
                                & \multicolumn{3}{c|}{Euclidean Cost} & \multicolumn{3}{c}{Random Cost} \\ \midrule
                                & $\hat{t}_{init}$  & $\hat{t}_{opt}$  & $\hat{t}_{term}$    & $\hat{t}_{init}$    & $\hat{t}_{opt}$  & $\hat{t}_{term}$         \\ \midrule
\arastar                        & 19 (0.923)         & 47      & 50          & 41 (0.902)             & 99       & 178               \\
\epase                          & 11            & 11      & 11          & 38             & 38       & 38          \\
\aepasenaive                    & 6  (0.948)         & 159     & 200          & 10 (0.951)              & 396      & 767           \\
\textbf{\aepase}                         & \textbf{6 (0.949)}         & \textbf{14}      & \textbf{16}          & \textbf{10 (0.954)}       & \textbf{27}       & \textbf{44}          \\ \bottomrule 
                                & $\hat{\speedup}_{init}$ & $\hat{\speedup}_{opt}$ & $\hat{\speedup}_{term}$ & $\hat{\speedup}_{init}$ & $\hat{\speedup}_{opt}$ & $\hat{\speedup}_{term}$ \\ \midrule
\arastar                        & 2.69                 & 3.62       & 2.89        & 3.52               & 3.70         & 3.88             \\
\epase                          & 1.75                 & 1.00       & 0.70        & 4.17               & 1.82         & 0.86             \\
\aepasenaive                    & 0.98                 & 9.19       & 11.94        & 1.01               & 13.12        & 16.44             \\ \bottomrule 
\end{tabular}
}
\caption{Top: Mean time ($\si{\milli\second}$) to find the initial feasible solution ($\hat{t}_{init}$), discover optimal solution ($\hat{t}_{opt}$) and prove optimal solution ($\hat{t}_{term}$). Numbers in parenthesis in the $\hat{t}_{init}$ columns are the initial optimality ratios. Bottom: Speedup of \aepase over the baselines.}
\label{aepase/tab/time_speedup}
\end{table}

Table~\ref{aepase/tab/time_speedup} top shows raw planning times for three stages. $\hat{t}_{init}$ is the mean time to generate the first solution, $\hat{t}_{opt}$ is the mean time to first discover the optimal solution in hindsight, and $\hat{t}_{term}$ is the mean time to provably generate the optimal solution by the final $\textsc{ImprovePath}$ call with $\wh=1$. Table~\ref{aepase/tab/time_speedup} bottom presents the average speedup of \aepase over the baselines (${t}_{baseline}/{t}_{\aepase}$). This is generated by computing the speedup for each run and then averaging them over all runs and all maps. \aepasenaive and \aepase compute the initial solution faster than \arastar due to parallelization and than \epase due to the high inflation on the heuristic. 
\aepase computes the provably optimal solution quicker than \aepasenaive and \arastar, but slower than \epase. This is expected since \epase is not an anytime algorithm and runs a single optimal search. However, \aepase can discover the optimal solution in hindsight faster than \epase in the random cost map. This means that even if the time budget runs out before the \aepase runs its final iteration with $\wh=1$ to provably generate the optimal plan and the robot executes the best plan so far, it may still end up behaving optimally. This is an important and useful empirical result for real-time robotics.





\begin{figure}[!htb]
    \centering
    \includegraphics[width=\columnwidth]{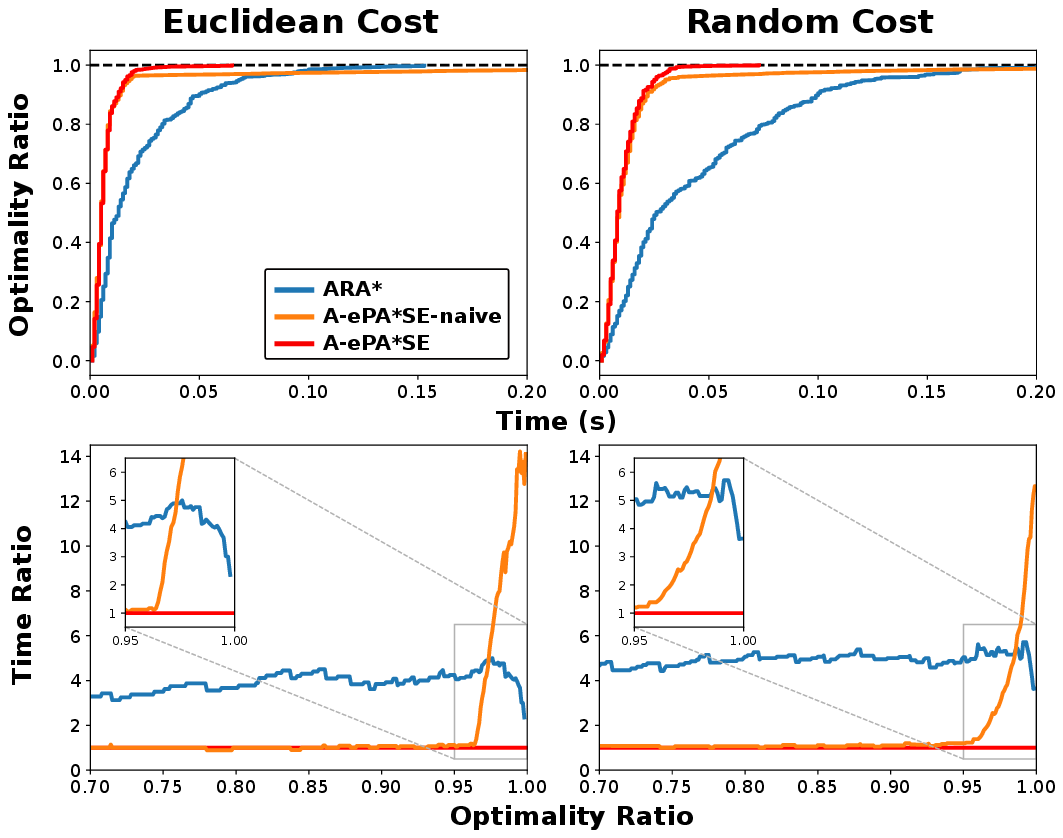}
    \caption{ Top: The mean optimality ratio an algorithm achieves at a specific time point. Bottom: Planning times of the baselines divided by that of \aepase to achieve the same mean optimality ratio. In other words, for a specific algorithm, the bottom plots show how much slower (y-axis) than \aepase that algorithm gets to a solution that has a specific optimality factor (x-axis). The plot in the 0.95-1.00 optimality range is zoomed into.}
    \label{aepase/fig/combined_time_speedup}
\end{figure}

Fig.~\ref{aepase/fig/combined_time_speedup} (top) shows the optimality ratio (optimal cost / actual cost) achieved by an anytime algorithm at a specific time point. For every problem, we calculate the optimality ratio at every time point when \textsc{ImprovePath} returns. We then discretize time and assign each time point with the best optimality ratio achieved so far. This is then averaged across all maps and problems separately for the two different cost maps. To show the relative performance, Fig.~\ref{aepase/fig/combined_time_speedup} (bottom) divides the time it takes the baselines to achieve a given optimality ratio by the time of \aepase to achieve the same optimality ratio. Specifically, the plot represents how  many times  slower an algorithm is than \aepase in computing a solution with a certain optimality factor represented by the x-axis. We see that \arastar takes significantly longer to reach the same optimality ratio as compared to \aepase. \aepasenaive does as good as \aepase for lower optimality ratios, but it takes significantly longer to achieve optimality because it does not reuse previous search effort. \aepase outperforms \arastar as predicted due to the efficiency gained from parallelization.

\subsubsection{Summary of results}
The experimental evaluation demonstrates the advantages of \aepase over the baselines. 
\begin{itemize}
    \item Compared to \arastar, both Fig.~\ref{aepase/fig/combined_time_speedup} and Table.~\ref{aepase/tab/time_speedup} indicate that \aepase outperforms \arastar in planning time.
    
    \item  As shown in Table~\ref{aepase/tab/time_speedup}, \aepase and \aepasenaive both find the initial solution at around 0.95 optimality. This implies that, on average, the solution cost improvement primarily happens in the 0.95-1.0 optimality range, making it the range of interest for analysis. Fig.~\ref{aepase/fig/combined_time_speedup} (zoomed in) shows that \aepase improves the optimality ratio quicker than \aepasenaive in the that range region.
    
    \item Compared to \epase, \aepase has an anytime behavior where it quickly computes a feasible solution and then improves it over time. Additionally, it computes the optimal solution in hindsight ($\hat{t}_{opt}$) faster than \epase in the random cost map, which is a useful insight in the real-time robotics context.
\end{itemize}

%% file: 08conclusion_future.tex
In this work, we present an anytime parallelized search algorithm \aepase. Our experiments demonstrated that \aepase achieves a significant speedup over ARA* in both computing an initial solution and then improving it to compute the optimal solution. Additionally, the anytime property of \aepase makes it potentially more useful than \epase in a range of real-time robotics domains. In the current formulation of \aepase, both the initial heuristic inflation $\wh_0$ and the decrement $\Delta\wh$ between successive \improve calls are parameters to be tuned. In the future, \aepase can be extended to a non-parametric formulation.

%% file: 09acknowledgements.tex
This work was supported by the ARL-sponsored A2I2 program, contract W911NF-18-2-0218, and ONR grant N00014-18-1-2775.